\documentclass{article}

\PassOptionsToPackage{numbers,square,comma,sort&compress}{natbib}

 \usepackage[dblblindworkshop, final]{neurips_2025}

\usepackage[utf8]{inputenc} 
\usepackage[T1]{fontenc}    
\usepackage{hyperref}       
\usepackage{url}            
\usepackage{booktabs}       
\usepackage{amsfonts}       
\usepackage{nicefrac}       
\usepackage{microtype}      
\usepackage{xcolor}         

\usepackage{lineno}
\usepackage{amsmath}
\usepackage{graphicx}
\usepackage[table]{xcolor}
\usepackage[most]{tcolorbox}

\definecolor{darkblue}{rgb}{0, 0, 0.5}
\hypersetup{colorlinks=true, citecolor=darkblue, linkcolor=darkblue, urlcolor=darkblue}

\definecolor{myBlue}{rgb}{0.176, 0.521, 0.773}
\definecolor{myRed}{rgb}{0.8, 0.215, 0.145}
\definecolor{myLightBlue}{rgb}{0.38, 0.71, 0.94}
\definecolor{myGrey}{rgb}{0.73, 0.73, 0.74}

\title{It's LIT!\\Reliability-Optimized LLMs with Inspectable Tools}
\workshoptitle{Workshop on Multi-Turn Interactions in Large Language Models}

\author{%
  Ruixin Zhang \\
  Department of Computer Science\\
  Duke University\\
  \texttt{ruixin.zhang@duke.edu} \\
  \And
  Jon Donnelly \\
  Department of Computer Science\\
  Duke University\\
  \texttt{jon.donnelly@duke.edu} \\
  \And
  Zhicheng Guo \\
  Department of Computer Science\\
  Duke University\\
  \texttt{zhicheng.guo@duke.edu} \\
  \And
  Ghazal Khalighinejad \\
  Department of Computer Science\\
  Duke University\\
  \texttt{ghazal.khalighinejad@duke.edu} \\
  \And
  Haiyang Huang \\
  Department of Computer Science\\
  Duke University\\
  \texttt{haiyang.huang@duke.edu} \\
  \And
  Alina Jade Barnett \\
  Department of Computer Science\\
  Duke University\\
  \texttt{alina.barnett@duke.edu} \\
  \And
  Cynthia Rudin \\
  Department of Computer Science\\
  Duke University\\
  \texttt{cynthia@cs.duke.edu}
}

\begin{document}

\maketitle

\begin{abstract}
Large language models (LLMs) have exhibited remarkable capabilities across various domains. The ability to call external tools further expands their capability to handle real-world tasks. However, LLMs often follow an opaque reasoning process, which limits their usefulness in high-stakes domains where solutions need to be trustworthy to end users. LLMs can choose solutions that are unreliable and difficult to troubleshoot, even if better options are available.
We address this issue by forcing LLMs to use external -- more reliable -- tools to solve problems when possible. We present a framework built on the tool-calling capabilities of existing LLMs to enable them to select the most reliable and easy-to-troubleshoot solution path, which may involve multiple sequential tool calls. 
We refer to this framework as LIT (LLMs with Inspectable Tools). In order to support LIT, we introduce a new and challenging benchmark dataset of 1,300 questions and a customizable set of reliability cost functions associated with a collection of specialized tools. These cost functions summarize how reliable each tool is and how easy it is to troubleshoot. For instance, a calculator is reliable across domains, whereas a linear prediction model is not reliable if there is distribution shift, but it is easy to troubleshoot. A tool that constructs a random forest is neither reliable nor easy to troubleshoot.
These tools interact with the Harvard USPTO Patent Dataset and a new dataset of NeurIPS 2023 papers to solve mathematical, coding, and modeling problems of varying difficulty levels. We demonstrate that LLMs can achieve more reliable and informed problem-solving while maintaining task performance using our framework. 
\end{abstract}

\begin{figure}[b]
    \centering
    \includegraphics[width=0.7\linewidth]{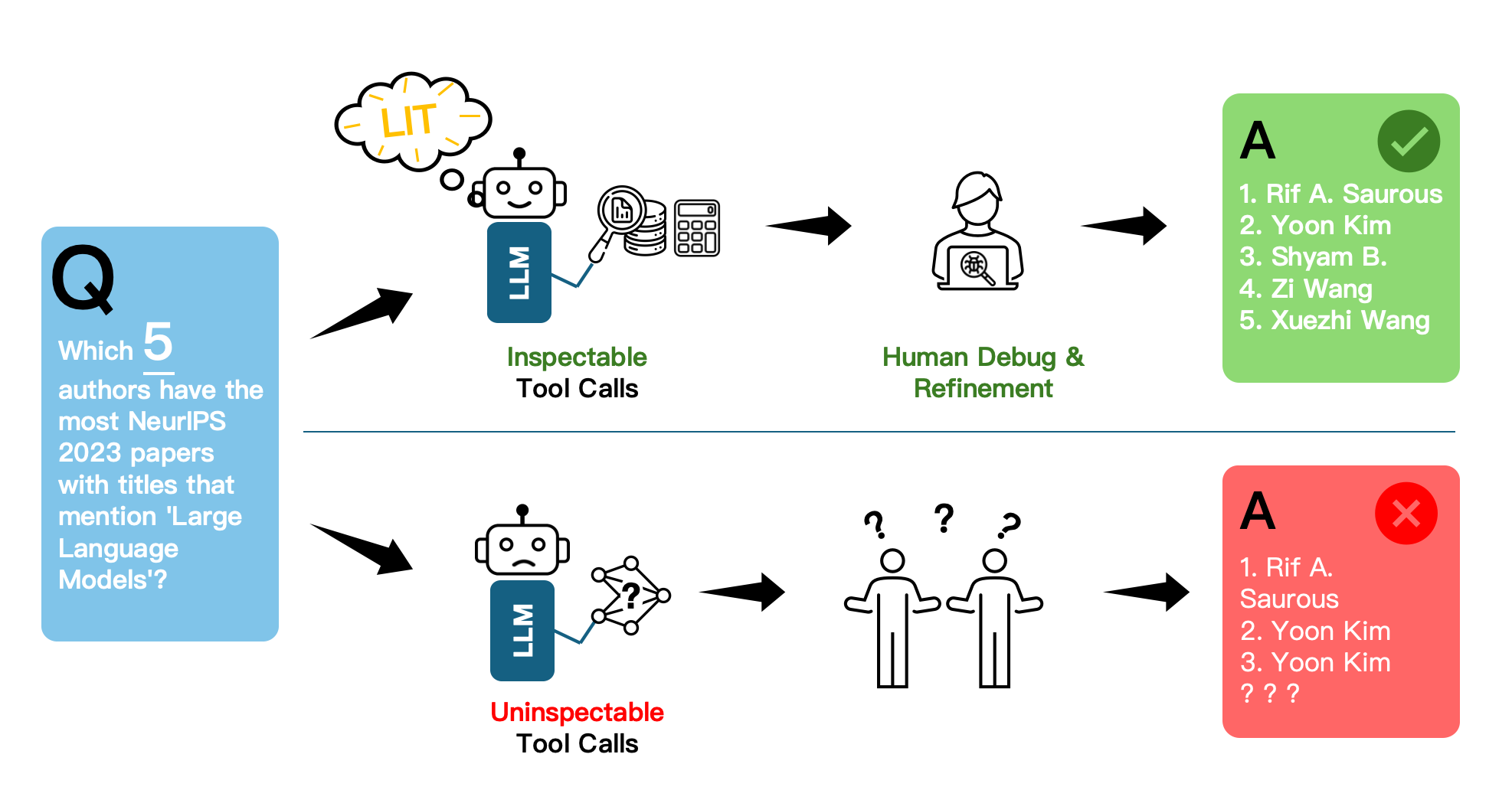}
    \caption{Conceptual summary. By using the LIT framework, the LLM selects more reliable and inspectable tools, allowing human users to debug and refine the solution proposed by the LLM, producing the correct result. In contrast, the vanilla LLM without LIT selects uninspectable solutions with reasoning that cannot be inspected or corrected.}
    \label{fig:overview}
\end{figure}


\section{Introduction}
Large language models (LLMs) are arguably the most complicated black box machine learning algorithms currently in existence. Their transformer architectures make them extremely difficult to trust and troubleshoot. When they answer difficult questions wrong, no one knows why, and no one can figure out why \citep{Farquhar2024}.
One path towards a potential solution is the use of \textit{external tools} by LLMs \citep{schick2023toolformer}. 
Rather than depending on a single huge LLM to solve every problem or perform every calculation, external tools specialize in various tasks and can be called by the LLM to perform them as part of solving a larger problem. For instance, rather than depending on a single LLM to solve (502*2952+323)/63, the LLM could call a calculator. Other tools might retrieve datasets, execute code, or perform other tasks. 
Allowing LLMs to call external tools offers a promising way to mitigate their complexity for several reasons. First, tools are specialized and more reliable; second, using them makes the LLM's logic more modular, and allows developers to take responsibility for key steps in forming a solution \citep{belcak2025small, gao2023pal, gou2023tora, yuan2023craft, schick2023toolformer, shen2024hugginggpt}.

Importantly, not all tools are created the same. They vary tremendously in their reliability and ease of troubleshooting. For instance, while a calculator is completely reliable given the correct inputs, a specialized model for solving physics problems may not be (though it might be more reliable for physics problems than the more general LLM calling it, and certainly easier to troubleshoot). Prior work on tool-based LLMs~\citep{qu2024tool, wang24colm} failed to account for the relative reliability and inspectability of tools within their toolboxes, which means their LLMs' solutions are likely more opaque and difficult to troubleshoot than necessary. Prior to the current work, there existed no metrics, baseline methods, standard frameworks, or question banks for exploring the reliability of tool selection in LLMs. 

We address this shortcoming by explicitly encouraging an LLM to prefer reliable and inspectable tools over alternatives by assigning a reliability and inspectability cost to each tool call and choosing the solution with the minimum cost. This allows the LLM to provide reliable and inspectable reasoning whenever possible without sacrificing the ability to answer difficult prompts correctly. This is illustrated in Figure \ref{fig:overview}. For example, imagine that we asked an LLM which five authors have the most NeurIPS 2023 papers with titles that mention ``Large Language Models.'' The LLM might simply provide a random list of 5 author names, but we would prefer that it use SQL code with reliable logic to identify these 5 authors.

We introduce the LLMs with Inspectable Tools (LIT) framework. During inference, the LLM generates multiple candidate solutions, each consisting of a sequence of tool calls. For each solution, the model computes the total cost based on the individual costs of each tool that it uses. It then selects the most reliable and inspectable option that still ensures accurate results. Once the optimal solution is identified, the LLM sequentially executes the necessary tool calls to produce the final answer.

We evaluate this framework using a suite of specialized tools with varying levels of reliability and inspectability, a novel benchmark dataset of 1,300 questions with varying levels of difficulty, and two distinct problem domains. We use 50\% of these questions for validation, and the remaining 50\% for testing (the LIT framework does not include any model training). We find that prompting the LLM to prefer reliable and inspectable tools dramatically increases the transparency of the given solutions, and also improves the model's accuracy in answering most of these questions. In summary, we provide the following contributions:
\begin{enumerate}
    \item We introduce a prompting framework that explicitly accounts for differences in the reliability and inspectability of tools  (Appendix \ref{sec:prompt_details} contains detailed prompts). We show that, across several LLMs, this framework leads to responses that exhibit enhanced inspectability while maintaining accuracy.
    \item We provide a suite of 8 tools with varying levels of reliability and inspectability, including tools for arithmetic operations (Calculator), dataset retrieval (DBLoader), flexible code execution (PandasInterpreter, PythonInterpreter), time series forecasting (Forecaster), text classification (TextualClassifier), language model inference (LLMInferencer), and result consolidation (Finish).\footnote{Each tool is designed to offer deterministic results for reproducibility.} 
    \item We introduce a new benchmark for tool-based LLMs consisting of 1,300 questions about two external databases: the Harvard USPTO Patent Dataset \citep{suzgun2024harvard} and a novel database with metadata from NeurIPS papers from 2023.
\end{enumerate}

\section{Related work}

\subsection{Tool learning} Our work is closely related to the topic of tool learning. Tool learning allows LLMs to use external tools to solve complex problems, gain real-time domain knowledge, or mitigate their own limitations \citep{vemprala2024chatgpt, nakano2021webgpt, qin2023webcpm, wang2024executable, hao2024toolkengpt, tang2023toolalpaca, gou2023tora, yuan2023craft}. Although previous work has shown the effectiveness of tool learning, if there are multiple choices of viable tools available, the algorithm will somehow choose one set of tools (or choose to call none at all), and the decision logic would likely remain opaque to human operators. In previous work on both single-step problem solving \citep{schick2023toolformer, shen2024hugginggpt, lu2024chameleon} and iterative multi-step tool calling with feedback \citep{li2023api, song2306restgpt, qin2023toolllm, gao2024confucius}, the LLM is optimized \textit{only} for task success and solution correctness, while in reality, optimizing for other objectives, such as reliability or the ability to troubleshoot, may yield more useful results. In this work, we focus on reliability and inspectable tool usage, which is compatible with any of the above tool-calling paradigms. Our experiments examine the case of sequential, dependent tool calling. 

\subsection{Tool learning datasets and benchmarks} 
There has been a rapid development of benchmark datasets for tool learning with LLMs \citep{li2023api, patil2023gorilla, xu2023tool, tang2023toolalpaca, song2306restgpt, shen2023taskbench}. Existing benchmark datasets largely focus on public API calls as tools, and evaluations mainly focus on the call counts and metrics based on task success (e.g., precision, F-1 score, rank). Our dataset provides a much-needed supplement to existing benchmarks, allowing the manual creation of additional tools and customizable reliability metrics in addition to traditional success rate-based metrics. Its built-in tools range from custom functions to external models. The questions in our benchmark dataset have varying degrees of difficulty, enabling researchers to measure the ability of LLMs to select reliable and inspectable tools.

\section{Framework}

Given a set of tools, LIT consists of two key components: a set of tool-cost functions, and a carefully 
designed few-shot
prompt to guide the underlying LLM to prefer more reliable and inspectable -- or, lower cost -- tools. We first provide the general principles for defining tool-cost functions, then introduce the prompt strategy used in LIT.

\subsection{Tool cost design}
\label{subsec:cost_design}
The cost of a solution is the combined costs of all the tools used in the solution sequence. 
%
Drawing inspiration from human-computer interaction (HCI) literature, which emphasizes reliability, debuggability, and simplicity as desirable qualities of systems \cite{raees2024explainable}, we consider three primary criteria when determining the cost of a tool: whether the tool achieves robust performance across inputs, the ease of debugging the tool, and the complexity of the arguments to the tool. These costs can be determined differently by the user in each context: while a calculator may be reliable across domains, a specialized LLM model may be less reliable when used out of its training distribution.
%
We present a list of reasonable values in Table \ref{tab:cost_component}, applying each criterion to the tools considered, although these values can easily be adjusted and set according to the problem context. Each criterion is described in detail below:

\textbf{Costs should consider how robust performance is across inputs (P).} We prefer tools that, given some input, reliably produce exactly the output a user expects. For example, tools like DBLoader and the Calculator demonstrate robust performance because DBLoader simply loads a dataset and any reasonable calculator always produces correct results for arithmetic operations. 
In contrast, model-based tools exhibit less robust performance. For example, the Forecaster (based on an AutoRegressive Integrated Moving Average -- ARIMA -- model) relies on the stationarity and linearity of the data, sufficient data points, and the presence of autocorrelation \citep{box2015time}. ARIMA's predictions are unreliable without these conditions. 

\textbf{Costs should consider ease of debugging (D).} We consider tools that can easily be debugged. For example, tools like the PandasInterpreter and PythonInterpreter are designed to accept code as their arguments. This trait makes them easier to debug, as modifying the input arguments to address errors or adjust behavior is straightforward, since the mapping between a given input argument and the tool's output is completely transparent. 
In contrast, tools like the Forecaster and TextualClassifier have more rigid argument structures. A Forecaster typically requires previous time series data and a forecast length as its primary inputs, while a TextualClassifier 
calls a trained model for predicting a binary outcome, which could be either a BERT model \cite{devlin-etal-2019-bert}, a logistic regression model, or a convolutional neural network model. The LLM decides which arguments to put in, what inputs to use, which model type to use, and which target variable to predict.

The limited number of argument combinations available for these tools reduces their flexibility. It can also make troubleshooting more difficult, as potential sources of errors are harder to modify. Moreover, it is not always clear how changing one parameter of a predictive model will change its predictions, particularly for complex predictive models.

\textbf{Costs should consider the complexity of arguments (C).} We consider tools to be more reliable and easier to troubleshoot when they take simple arguments. 
The susceptibility of tools to errors often correlates with the complexity of the arguments they handle. For instance, tools like the PandasInterpreter and PythonInterpreter become more prone to errors as the complexity of the input code increases. Factors such as additional lines of code or the inclusion of multiple imported packages introduce more dependencies and interactions, raising the likelihood of issues \citep{trisovic2022largescale}. Similarly, in the case of the TextualClassifier, the choice of trained model significantly influences error susceptibility. More complex models, such as BERT, involve additional parameters and intricate processes, introducing additional potential points of failure compared to simpler models like logistic regression. Consequently, tools handling more complex arguments incur higher costs due to the increased risk of execution issues.

\subsection{Reliability prompt and LLM reasoning}
\begin{figure}
    \centering
    \includegraphics[width=0.95\linewidth]{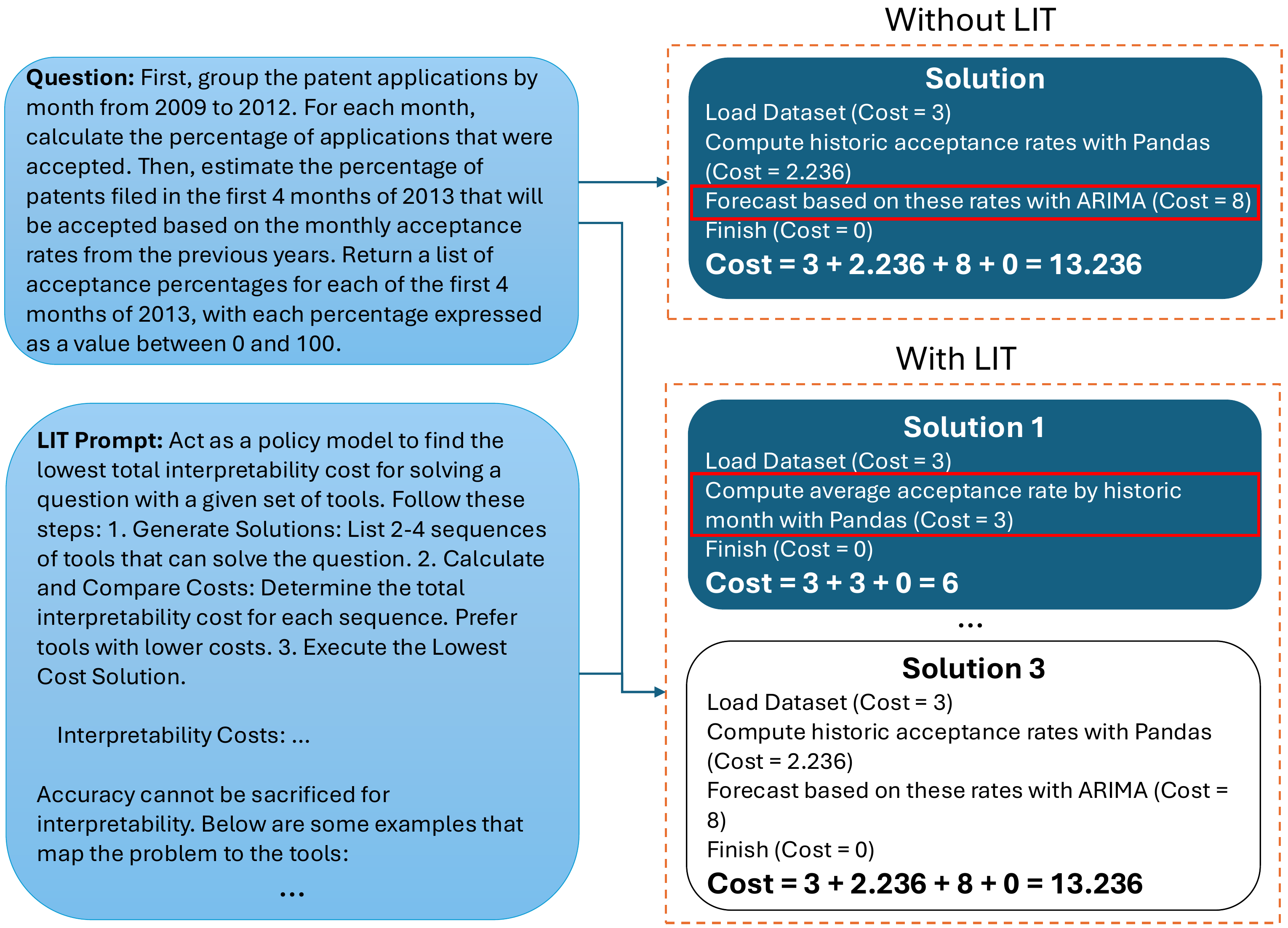}
    \caption{A simplified example of the prompting framework used by LIT. In LIT, we provide the model a cost for each tool and instruct the model to provide multiple alternative solutions, selecting the one with the lowest cost when possible.}
    \label{fig:prompt_example}
\end{figure}

We now introduce the prompt strategy used in LIT. We designed a prompt (described in detail in Appendix \ref{sec:prompt_details}) that incorporates customizable cost formulas for each tool, as described in the previous section, along with 5 examples of reliable and inspectable solutions for questions not included in our benchmark dataset. 
Within this prompt, the LLM is instructed to generate as many solutions as possible, up to a maximum of four, to avoid duplication and manage token usage efficiently. Each solution involves sequential tool calls. The LLM independently generates reasoning chains, which include calculating the overall cost for each solution. It then compares these costs to select the most reliable/inspectable solution while ensuring that accuracy is not compromised. Part of the LLM reasoning is also to decide how long each sequence of tool calls should be. The sequence is ended by calling the tool ``Finish.'' Once the optimal solution is chosen, the LLM proceeds to execute it, making tool calls sequentially. Figure \ref{fig:prompt_example} provides a simple example of this process.

\section{Benchmark details}
Let us describe the new reliability/inspectability benchmark. We created a set of 1,300 questions concerning one existing and one new dataset, along with a suite of 8 tools that each LLM can leverage.

\subsection{Datasets}
\label{sec:exp_datasets}
\paragraph{Harvard USPTO Patent Dataset (HUPD)} The HUPD dataset \citep{suzgun2024harvard} comprises 4.5 million patent applications filed with the United States Patent and Trademark Office (USPTO) between 2004 and 2018. Each patent application is structured as a JSON file that contains the abstract, background, claims, summary, and full description. In addition to the text fields, the JSON file includes metadata such as the inventors' names, cities, filing dates, and the USPTO's final decision. 

The abundance of metadata allows us to use various tools that can be called by an LLM.
Our questions can be answered with either reliable/inspectable or unreliable/uninspectable tools. For example, we ask the LLM to estimate the average filing time within a certain time frame, or to predict the likelihood of an application being accepted or not. These problems could be handled by the LLM alone, or by calling tools that make database queries, calling tools that constrain the selection of the database to given values, tools that create datasets for logistic regression and then run it, etc. 

\paragraph{NeurIPS 2023 Papers Dataset}
Using web scraping techniques, we collected information for each accepted NeurIPS 2023 paper, including the title, author list, abstract, topic, and whether the paper was accepted for an oral presentation. For this dataset, we might analyze the distribution of the number of authors or estimate the likelihood of a paper being selected for an oral presentation.


\subsection{Tools}
\label{subsec:tool_definitions}
We developed a suite of 8 tools with varying levels of reliability/inspectability for the LIT framework, described here and in Appendix \ref{sec:modelbasedtools_details}. This suite ranges from simple, reliable tools such as a basic calculator and a database loader to more complex, uninspectable modeling tools that employ BERT-based and convolutional neural network models. These tools enable LLMs to effectively address queries related to each dataset. In our experiments, we consider sequential, dependent tool calling.

-  Calculator: This tool performs arithmetic operations based on input expressions containing only numbers and operators (e.g., 2×3).\\
- DBLoader: This tool loads a specified database (either the HUPD or the NeurIPS 2023 Papers Dataset) as indicated by the DBName parameter. It also allows filtering of the dataset by specifying a subset, which can include either years or specific rows of the dataframe.\\
-  PandasInterpreter: This tool interprets Pandas code written in Python, utilizing the database stored in the variable \textit{df}, and returns a dictionary containing the values of variables defined within that code. It is particularly useful when the inquiry necessitates data manipulation on structured databases.\\
-  PythonInterpreter: This tool interprets Python code and returns a dictionary containing the values of variables defined within that code.\\
-  Forecaster: This tool executes a specified forecasting model (either an ARIMA model or a linear regression model) on historical data to predict subsequent data points (e.g., the patent acceptance rates over the next three months).\\
-  TextualClassifier: This tool applies a trained binary classification model (either BERT, logistic regression, or convolutional neural network) to an input string to make predictions. The model, input, and target variable must be specified.\\
-  LLMInferencer: This tool leverages the current LLM to generate solutions only when answers cannot be determined through other tools or when the queries involve complex reasoning.\\
-  Finish: This tool concludes the task and returns the final answer, ensuring that all variable values are derived directly from the output of the previous tool call and are of the correct return data type.

\subsection{Questions}
The questions are generated from carefully designed templates that require the use of multiple tools in sequence to solve the problem. No problem can be addressed with a single tool. We define 13 question templates, provided in Appendix \ref{sec:question_details}, which are used to produce many distinct questions by filling in different specific values. For example, the question template ``How does the number of patent applications filed in \{year1\} compare proportionally to those filed in \{year2\}?'' may produce questions like ``How does the number of patent applications filed in 2019 compare proportionally to those filed in 2020?'' or ``How does the number of patent applications filed in 2018 compare proportionally to those filed in 2020?'' Each question has multiple possible solution approaches and a singular answer, so that under prompts specifying costs, the LLM generates several solution sets and selects the one with the lowest cost.

All questions require accessing information from the provided databases and are classified into three categories: easy, medium, and hard. Easy questions are designed to be optimally solved using reliable/inspectable tools, e.g., ``How does the number of patent applications filed in \{year1\} compare proportionally to those filed in \{year2\}?'' Medium questions can be solved effectively using either reliable/inspectable or unreliable/uninspectable tools, with comparable performance, e.g., ``For a patent application not present in the database, with an abstract \{abstract\_content\}, predict whether it will be accepted. Return either `ACCEPTED' or `NOT ACCEPTED.''' Hard questions are structured to be best addressed by black box tools, e.g., ``Predict the best-fit topic for the title of a NeurIPS paper, not present in the database: \{title\}. Options: \{topic1\}, \{topic2\}, \{topic3\}.''

\section{Experimental results}
\begin{table*}[t]
    \centering
    \resizebox{\textwidth}{!}{
        \begin{tabular}{c|c|c|c|c|c|c|c|c|c|c|c|c|c!{\vrule width 2pt}c|c|c}
            LLM & Q1 & Q2 & Q3 & Q4 & Q5 & Q6 & Q7 & Q8 & Q9 & Q10 & Q11 & Q12 & Q13 & easy & med & hard \\
            \hline
            GPT-3.5 & \cellcolor{myBlue} 5.40 & \cellcolor{myLightBlue} 4.79 & \cellcolor{myLightBlue}5.88 & \cellcolor{myLightBlue} 4.74 & \cellcolor{myLightBlue} 4.18 & \cellcolor{myBlue} 9.87 & \cellcolor{myGrey} NA & \cellcolor{myBlue} 18.70 & \cellcolor{myGrey} 30.00 & \cellcolor{myBlue} 15.67 & \cellcolor{myBlue} 30.00 & \cellcolor{myBlue} 30.00 & \cellcolor{myLightBlue} 24.86 & \cellcolor{myBlue} 5.81 & \cellcolor{myBlue} 17.32 & \cellcolor{myBlue} 28.29 \\
            GPT-3.5 LIT & \cellcolor{myBlue} 4.53 & \cellcolor{myLightBlue} 4.05 & \cellcolor{myLightBlue} 6.21 & \cellcolor{myLightBlue} 4.61 & \cellcolor{myLightBlue} 4.75 & \cellcolor{myBlue} 4.27 & \cellcolor{myGrey} 10.90 & \cellcolor{myBlue} 9.17 & \cellcolor{myGrey} NA & \cellcolor{myBlue} 11.30 & \cellcolor{myBlue} 4.87 & \cellcolor{myBlue} 18.57 & \cellcolor{myLightBlue} 26.07 & \cellcolor{myBlue} 4.74 & \cellcolor{myBlue} 10.46 & \cellcolor{myBlue} 16.50 \\
            GPT-4 & \cellcolor{myBlue} 5.91 & \cellcolor{myRed} 4.56 & \cellcolor{myBlue} 6.88 & \cellcolor{myBlue} 5.14 & \cellcolor{myLightBlue} 6.27 & \cellcolor{myBlue} 5.42 & \cellcolor{myBlue} 30.48 & \cellcolor{myBlue} 20.00 & \cellcolor{myLightBlue} 30.00 & \cellcolor{myBlue} 19.74 & \cellcolor{myRed} 30.00 & \cellcolor{myLightBlue} 30.00 & \cellcolor{myLightBlue} 30.00 & \cellcolor{myBlue} 5.70 & \cellcolor{myBlue} 25.06 & \cellcolor{myRed} 30.00 \\
            GPT-4 LIT & \cellcolor{myBlue} 5.29 & \cellcolor{myRed} 4.71 & \cellcolor{myBlue} 5.35 & \cellcolor{myBlue} 5.01 & \cellcolor{myLightBlue} 5.79 & \cellcolor{myBlue} 4.89 & \cellcolor{myBlue} 8.68 & \cellcolor{myBlue} 8.38 & \cellcolor{myLightBlue} 30.00 & \cellcolor{myBlue} 7.00 & \cellcolor{myRed} 32.58 & \cellcolor{myLightBlue} 30.00 & \cellcolor{myLightBlue} 30.00 & \cellcolor{myBlue} 5.17 & \cellcolor{myBlue} 13.52 & \cellcolor{myRed} 30.86 \\
            Gemini & \cellcolor{myLightBlue} 4.00 & \cellcolor{myLightBlue} 4.66 & \cellcolor{myBlue} 5.87 & \cellcolor{myLightBlue} 4.12 & \cellcolor{myLightBlue} 4.65 & \cellcolor{myBlue} 4.24 & \cellcolor{myBlue} 16.68 & \cellcolor{myLightBlue} 19.90 & \cellcolor{myLightBlue} 30.00 & \cellcolor{myBlue} 28.11 & \cellcolor{myLightBlue} 30.00 & \cellcolor{myLightBlue} 30.00 & \cellcolor{myLightBlue} 30.00 & \cellcolor{myLightBlue} 4.59 & \cellcolor{myBlue} 23.67 & \cellcolor{myLightBlue} 30.00 \\
            Gemini LIT & \cellcolor{myLightBlue} 4.00 & \cellcolor{myLightBlue} 5.90 & \cellcolor{myBlue} 4.05 & \cellcolor{myLightBlue} 4.00 & \cellcolor{myLightBlue} 4.52 & \cellcolor{myBlue} 3.91 & \cellcolor{myBlue} 4.11 & \cellcolor{myLightBlue} 17.33 & \cellcolor{myLightBlue} 30.00 & \cellcolor{myBlue} 11.58 & \cellcolor{myLightBlue} 30.00 & \cellcolor{myLightBlue} 30.00 & \cellcolor{myLightBlue} 30.00 & \cellcolor{myLightBlue} 4.40 & \cellcolor{myBlue} 15.76 & \cellcolor{myLightBlue} 30.00 \\
            Claude & \cellcolor{myBlue} 5.46 & \cellcolor{myBlue} 5.17 & \cellcolor{myBlue} 5.43 & \cellcolor{myBlue} 6.11 & \cellcolor{myBlue} 6.90 & \cellcolor{myBlue} 5.07 & \cellcolor{myLightBlue} 9.69 & \cellcolor{myBlue} 20.00 & \cellcolor{myLightBlue} 31.47 & \cellcolor{myBlue} 20.00 & \cellcolor{myLightBlue} 30.06 & \cellcolor{myLightBlue} 30.00 & \cellcolor{myLightBlue} 30.12 & \cellcolor{myBlue} 5.52 & \cellcolor{myBlue} 20.29 & \cellcolor{myLightBlue} 30.06 \\
            Claude LIT & \cellcolor{myBlue} 4.54 & \cellcolor{myBlue} 4.75 & \cellcolor{myBlue} 4.97 & \cellcolor{myBlue} 4.86 & \cellcolor{myBlue} 5.60 & \cellcolor{myBlue} 4.61 & \cellcolor{myLightBlue} 12.25 & \cellcolor{myBlue} 11.58 & \cellcolor{myLightBlue} 30.31 & \cellcolor{myBlue} 16.32 & \cellcolor{myLightBlue} 29.67 & \cellcolor{myLightBlue} 30.00 & \cellcolor{myLightBlue} 30.06 & \cellcolor{myBlue} 5.06 & \cellcolor{myBlue} 17.62 & \cellcolor{myLightBlue} 29.91 \\
            Llama-3.1 & \cellcolor{myBlue} 5.75 & \cellcolor{myLightBlue} 5.50 & \cellcolor{myLightBlue} 5.31 & \cellcolor{myLightBlue} 4.91 & \cellcolor{myLightBlue} 6.67 & \cellcolor{myLightBlue} 4.79 & \cellcolor{myLightBlue} 5.31 & \cellcolor{myLightBlue} 7.00 & \cellcolor{myLightBlue} 30.00 & \cellcolor{myBlue} 10.52 & \cellcolor{myLightBlue} 9.88 & \cellcolor{myLightBlue} 19.50 & \cellcolor{myBlue} 23.34 & \cellcolor{myLightBlue} 5.66 & \cellcolor{myLightBlue} 13.21 & \cellcolor{myBlue} 17.57 \\
            Llama-3.1 LIT & \cellcolor{myBlue} 4.25 & \cellcolor{myLightBlue} 5.35 & \cellcolor{myLightBlue} 5.29 & \cellcolor{myLightBlue} 5.04 & \cellcolor{myLightBlue} 6.32 & \cellcolor{myLightBlue} 4.89 & \cellcolor{myLightBlue} 6.22 & \cellcolor{myLightBlue} 7.00 & \cellcolor{myLightBlue} 30.00 & \cellcolor{myBlue} 7.43 & \cellcolor{myLightBlue} 8.24 & \cellcolor{myLightBlue} 19.54 & \cellcolor{myBlue} 8.93 & \cellcolor{myLightBlue} 5.19 & \cellcolor{myLightBlue} 12.66 & \cellcolor{myBlue} 12.24 \\
        \end{tabular}
    }
    \includegraphics[width=1.0\linewidth]{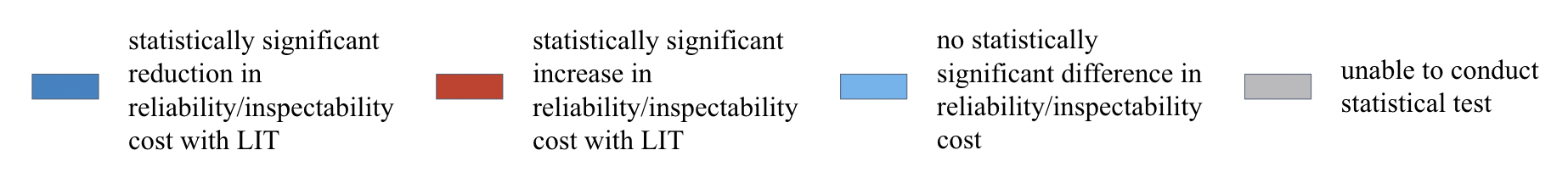}
    \caption{Comparison of reliability/inspectability cost. For each question (columns 1-13) and different LLMs (rows), we compare the mean cost of our LIT method with that of the standard, uninspectable baseline using a one-sided independent two-sample t-test. Columns 14-16 contain the averages for ``easy'' (Q1-Q6), ``medium'' (Q7-Q10) and ``hard'' (Q11-Q13) questions.}
    \label{tab:compare_cost}
\end{table*}

\begin{table*}
    \centering
    \resizebox{\textwidth}{!}{
        \begin{tabular}{c|c|c|c|c|c|c|c|c|c|c|c|c|c!{\vrule width 2pt}c|c|c}
            LLM & Q1 & Q2 & Q3 & Q4 & Q5 & Q6 & Q7 & Q8 & Q9 & Q10 & Q11 & Q12 & Q13 & easy & med & hard \\
            \hline
            GPT-3.5 & \cellcolor{myRed} 0.95 & \cellcolor{myRed} 0.83 & \cellcolor{myBlue} 0.78 & \cellcolor{myLightBlue} 0.56 & \cellcolor{myBlue} 0.06 & \cellcolor{myBlue} 0.39 & \cellcolor{myGrey} NA & \cellcolor{myLightBlue} 0.42 & \cellcolor{myGrey} 0.53 & \cellcolor{myBlue} 0.35 & \cellcolor{myRed} 0.30 & \cellcolor{myRed} 0.26 & \cellcolor{myLightBlue} 0.14 & \cellcolor{myLightBlue} 0.60 & \cellcolor{myBlue} 0.33 & \cellcolor{myRed} 0.23 \\
            GPT-3.5 LIT & \cellcolor{myRed} 0.61 & \cellcolor{myRed} 0.71 & \cellcolor{myBlue} 0.93 & \cellcolor{myLightBlue} 0.46 & \cellcolor{myBlue} \cellcolor{myBlue} 0.18 & \cellcolor{myBlue} 0.81 & \cellcolor{myGrey} 0.00 & \cellcolor{myLightBlue} 0.42 & \cellcolor{myGrey} NA & \cellcolor{myBlue} 0.56 & \cellcolor{myRed} 0.18 & \cellcolor{myRed} 0.17 & \cellcolor{myLightBlue} 0.21 & \cellcolor{myLightBlue} 0.62 & \cellcolor{myBlue} 0.25 & \cellcolor{myRed} 0.19 \\
            GPT-4 & \cellcolor{myRed} 0.93 & \cellcolor{myBlue} 0.90 & \cellcolor{myLightBlue} 0.96 & \cellcolor{myLightBlue} 1.00 & \cellcolor{myLightBlue} 0.92 & \cellcolor{myBlue} 0.16 & \cellcolor{myLightBlue} 0.00 & \cellcolor{myRed} 0.43 & \cellcolor{myRed} 0.67 & \cellcolor{myBlue} 0.31 & \cellcolor{myBlue} 0.28 & \cellcolor{myBlue} 0.85 & \cellcolor{myLightBlue} 0.54 & \cellcolor{myBlue} 0.81 & \cellcolor{myRed} 0.35 & \cellcolor{myBlue} 0.56 \\
            GPT-4 LIT & \cellcolor{myRed} 0.75 & \cellcolor{myBlue} 0.99 & \cellcolor{myLightBlue} 0.96 & \cellcolor{myLightBlue} 1.00 & \cellcolor{myLightBlue} 0.90 & \cellcolor{myBlue} 0.81 & \cellcolor{myLightBlue} 0.00 & \cellcolor{myRed} 0.41 & \cellcolor{myRed} 0.25 & \cellcolor{myBlue} 0.56 & \cellcolor{myBlue} 0.30 & \cellcolor{myBlue} 0.94 & \cellcolor{myLightBlue} 0.50 & \cellcolor{myBlue} 0.90 & \cellcolor{myRed} 0.31 & \cellcolor{myBlue} 0.58 \\
            Gemini & \cellcolor{myLightBlue} 1.00 & \cellcolor{myBlue} 0.27 & \cellcolor{myLightBlue} 0.91 & \cellcolor{myLightBlue} 0.40 & \cellcolor{myLightBlue} 0.27 & \cellcolor{myLightBlue} 0.97 & \cellcolor{myLightBlue} 0.00 & \cellcolor{myRed} 0.37 & \cellcolor{myBlue} 0.49 & \cellcolor{myRed} 0.45 & \cellcolor{myBlue} 0.40 & \cellcolor{myRed} 0.96 & \cellcolor{myLightBlue} 0.33 & \cellcolor{myBlue} 0.64 & \cellcolor{myRed} 0.33 & \cellcolor{myBlue} 0.56 \\
            Gemini LIT & \cellcolor{myLightBlue} 1.00 & \cellcolor{myBlue} 0.57 & \cellcolor{myLightBlue} 0.98 & \cellcolor{myLightBlue} 0.56 & \cellcolor{myLightBlue} 0.24 & \cellcolor{myLightBlue} 0.90 & \cellcolor{myLightBlue} 0.00 & \cellcolor{myRed} 0.36 & \cellcolor{myBlue} 0.54 & \cellcolor{myRed} 0.34 & \cellcolor{myBlue} 0.50 & \cellcolor{myRed} 0.96 & \cellcolor{myLightBlue} 0.48 & \cellcolor{myBlue} 0.71 & \cellcolor{myRed} 0.31 & \cellcolor{myBlue} 0.65 \\
            Claude & \cellcolor{myLightBlue} 1.00 & \cellcolor{myLightBlue} 0.94 & \cellcolor{myLightBlue} 1.00 & \cellcolor{myBlue} 0.40 & \cellcolor{myLightBlue} 0.82 & \cellcolor{myLightBlue} 1.00 & \cellcolor{myLightBlue} 0.01 & \cellcolor{myRed} 0.43 & \cellcolor{myRed} 0.71 & \cellcolor{myLightBlue} 0.31 & \cellcolor{myLightBlue} 0.61 & \cellcolor{myLightBlue} 1.00 & \cellcolor{myLightBlue} 0.58 & \cellcolor{myBlue} 0.86 & \cellcolor{myRed} 0.37 & \cellcolor{myLightBlue} 0.73 \\
            Claude LIT & \cellcolor{myLightBlue} 1.00 & \cellcolor{myLightBlue} 0.92 & \cellcolor{myLightBlue} 1.00 & \cellcolor{myBlue} 1.00 & \cellcolor{myLightBlue} 0.78 & \cellcolor{myLightBlue} 1.00 & \cellcolor{myLightBlue} 0.02 & \cellcolor{myRed} 0.42 & \cellcolor{myRed} 0.66 & \cellcolor{myLightBlue} 0.31 & \cellcolor{myLightBlue} 0.61 & \cellcolor{myLightBlue} 1.00 & \cellcolor{myLightBlue} 0.56 & \cellcolor{myBlue} 0.95 & \cellcolor{myRed} 0.35 & \cellcolor{myLightBlue} 0.72 \\
            Llama-3.1 & \cellcolor{myLightBlue} 0.75 & \cellcolor{myLightBlue} 0.36 & \cellcolor{myLightBlue} 1.00 & \cellcolor{myLightBlue} 0.55 & \cellcolor{myRed} 0.52 & \cellcolor{myLightBlue} 0.67 & \cellcolor{myLightBlue} 0.00 & \cellcolor{myBlue} 0.21 & \cellcolor{myRed} 0.68 & \cellcolor{myBlue} 0.24 & \cellcolor{myRed} 0.06 & \cellcolor{myBlue} 0.36 & \cellcolor{myLightBlue} 0.24 & \cellcolor{myRed} 0.64 & \cellcolor{myRed} 0.28 & \cellcolor{myBlue} 0.22 \\
            Llama-3.1 LIT & \cellcolor{myLightBlue} 1.00 & \cellcolor{myLightBlue} 0.37 & \cellcolor{myLightBlue} 1.00 & \cellcolor{myLightBlue} 0.37 & \cellcolor{myRed} 0.05 & \cellcolor{myLightBlue} 0.66 & \cellcolor{myLightBlue} 0.00 & \cellcolor{myBlue} 0.28 & \cellcolor{myRed} 0.48 & \cellcolor{myBlue} 0.30 & \cellcolor{myRed} 0.00 & \cellcolor{myBlue} 0.64 & \cellcolor{myLightBlue} 0.19 & \cellcolor{myRed} 0.58 & \cellcolor{myRed} 0.27 & \cellcolor{myBlue} 0.28 \\
        \end{tabular}
    }
    \includegraphics[width=1.0\linewidth]{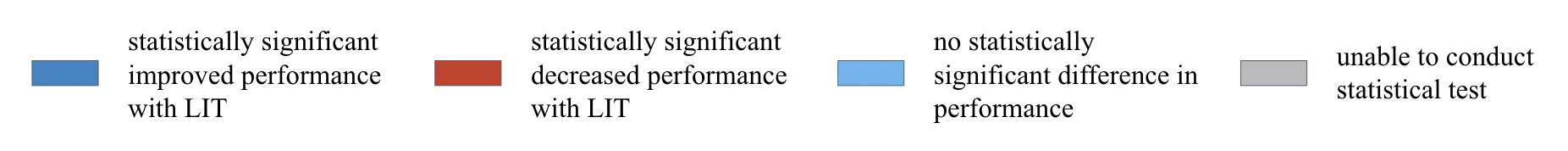}
    \caption{Comparison of performance (metrics are detailed in Appendix \ref{sec:metric_details} as they differ depending on the type of answer required). For each question (columns 1-13) and different LLMs (rows) we compare our LIT method to the standard, black box baseline using a one-sided independent two-sample t-test. Columns 14-16 contain the averages for ``easy'' (Q1-Q6), ``medium'' (Q7-Q10) and ``hard'' (Q11-Q13) questions. The LIT framework performed well for ``easy'' and ``hard'' questions, but performed worse on ``medium'' difficulty questions.}
    \label{tab:compare_perf}
\end{table*}

In this section, we evaluate the empirical efficacy of LIT. We consider 5 different LLMs with and without the LIT framework: GPT-3.5-Turbo \citep{gpt3}, GPT-4-Turbo \citep{achiam2023gpt}, Gemini-1.5-Pro-001 \citep{team2023gemini}, Claude-3.5-Sonnet-Latest \citep{claude}, and Meta-Llama-3.1-70B-Instruct-Turbo \citep{touvron2023llama}. We evaluate each model across all 13 question formats included in our benchmark, requiring access to each of the 8 tools and the 2 datasets included in the benchmark. Details of the evaluation metrics can be found in Appendix \ref{sec:metric_details}. Importantly, we always have access to the uninspectable baseline, so if LIT performs worse than it, we can always switch back to it. The question we investigate is whether LIT often has an advantage over the black box baseline.

\paragraph{LIT improves reliability/inspectability}
We evaluated the reliability/inspectability of LIT's answers by measuring the average cost (as specified in Table \ref{tab:cost_component}) of each model's solution across realizations of each question template. As shown in Table \ref{tab:compare_cost}, we found that LIT produced a solution with similar or better reliability/inspectability in {61} out of {65} cases. Notably, LIT improved reliability/inspectability for most questions across all 5 LLM backbones we considered. Moreover, we find that LIT improved or matched the average cost for easy- and medium-difficulty questions in every case except one (namely, Q2). We found that this trend did not hold as strongly for hard questions because LLMs produce high costs both with and without LIT, suggesting that these questions cannot be easily solved using the available reliable/inspectable tools.

\paragraph{LIT maintains performance}
We evaluated the performance of LIT for each question template. 
As shown in Table \ref{tab:compare_perf}, LIT resulted in improved or comparable performance in 48 out of 65 settings. 
That is, the gains in reliability and inspectability provided by LIT do not generally \textit{trade off} with performance; instead, they \textit{maintain or improve} performance. 

\begin{figure}[t]
    \centering
    \includegraphics[width=0.65\linewidth]{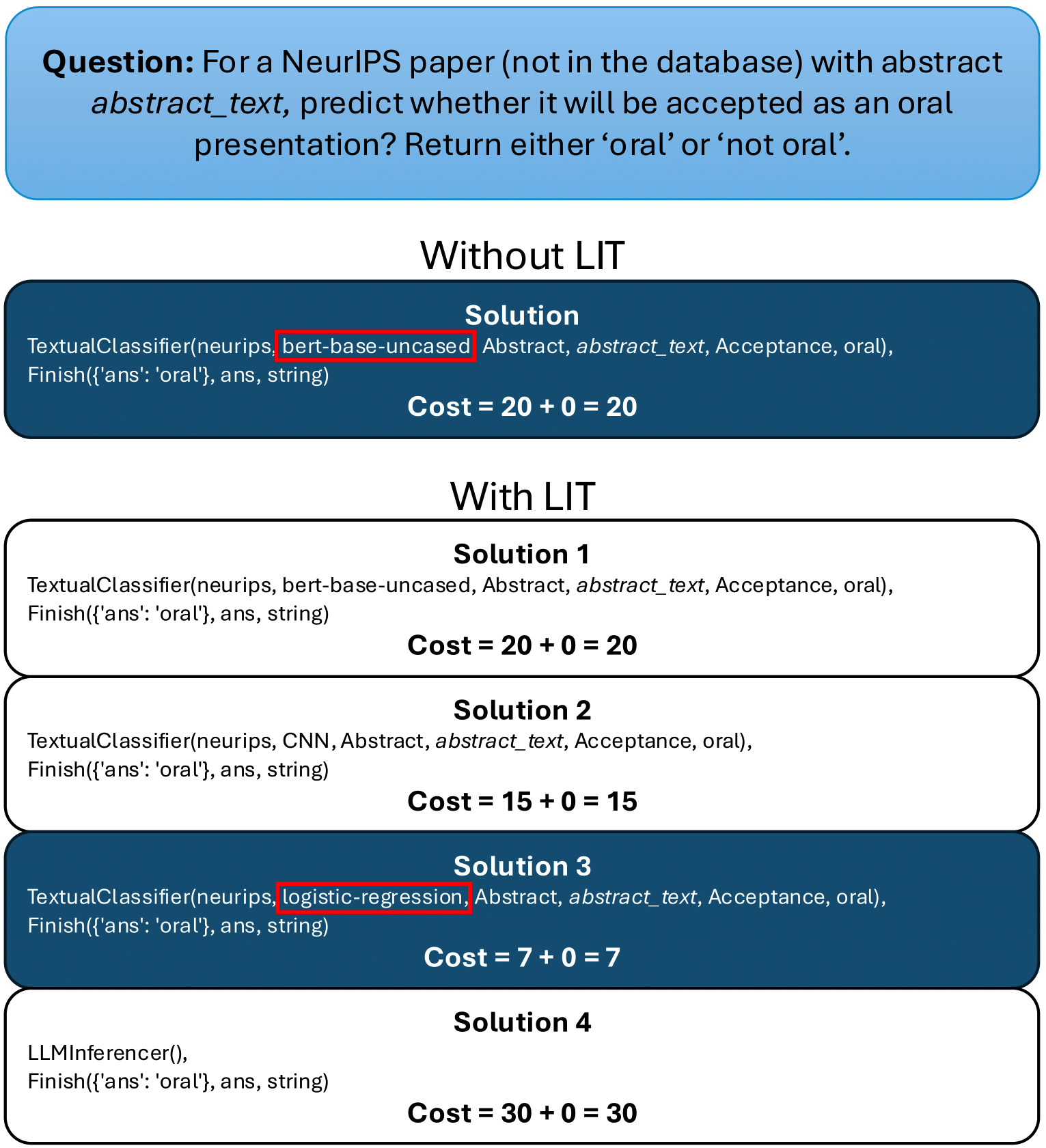}
    \caption{Solutions from an LLM to a question about whether a paper would be accepted to NeurIPS, with and without LIT. When using LIT, the model generates multiple candidate solutions and selects the one with the best cost. As a result, the model with LIT uses a logistic regression model rather than a BERT model to form its prediction, thereby using a substantially more inspectable tool. ``Finish'' tool signifies the end of logic stream and checks that the correct data type is returned. It has 0 cost. }
    \label{fig:prediction_example}
\end{figure}

\paragraph{LIT's responses are easier to debug}
Here, we present example responses from Claude-3.5-Sonnet-Latest with and without LIT. 
Figure \ref{fig:prediction_example} presents an illustrative case from Question 10, in which the model must predict whether a paper would be accepted for oral presentation at NeurIPS based on its abstract. In this case, the black-box solution uses the pre-trained BERT model within the TextualClassifier tool, while LIT's solution uses a pre-trained logistic regression model within this tool. The models achieve comparable accuracy, however, the logistic regression model offers a significant advantage. Its coefficients can be directly accessed and analyzed, providing clear insight into how it generates predictions, making it easy to troubleshoot. In contrast, the BERT model, with its vast number of parameters, has an inexplicable decision-making process. This is captured by the lower cost of 7 for the LIT solution compared to 20 for the black-box solution. Therefore, while both solutions are equally accurate, the LIT solution provides a distinct advantage for understanding and debugging the outputs. Appendix \ref{sec:additional_example} provides an additional example.


\paragraph{Our benchmark is challenging}
While LIT demonstrates improved reliability/inspectability and comparable performance to unaltered LLMs, we found that a variety of questions in this benchmark are not well solved by \textit{any} existing LLM techniques. As shown in Table \ref{tab:compare_perf}, no model achieves strong performance on Question 7 (with $R^2$ scores close to 0, indicating a failure to explain the variability in the target variable and performance only marginally better than that of a na\"ive mean predictor), and performance is generally low for all medium and hard questions. This is particularly notable because medium-difficulty questions tend to be solvable using multiple different tools, meaning many different solutions with different reliability/inspectability costs are possible if an LLM can find them. As such, this represents a direct avenue for future work.

\section{Conclusion}
We introduced LIT, a framework for reliable and inspectable tool use for LLMs. In order to evaluate LIT, we introduced a novel benchmark of 1,300 questions and a suite of 8 tools accessing two distinct external datasets. We showed that, simply by applying LIT, we can steer a wide variety of modern LLM models to select more reliable and inspectable tools without substantial cost to task performance.

LIT represents a substantial step towards more trustworthy LLM deployment. By leading LLMs to reason with inspectable tools, LIT provides a new level of transparency in LLM reasoning to users. Moreover, the benchmark introduced in this work will enable further work in this direction, ultimately leading to a more reliable future for LLMs.

We are aware of one primary limitation. 
The prompting strategy used in LIT increases both the amount of tokens given as input to each LLM and the amount of tokens generated as output (since the LLM must compare multiple solutions). This may pose a challenge when users are constrained by context limits. Future work should aim to improve the token efficiency of LIT.

\section*{Acknowledgments}
We acknowledge funding from the National Science Foundation under 
grants HRD-2222336. Additionally, this material is based upon work supported by the National Science
Foundation Graduate Research Fellowship under Grant No. DGE 2139754.


\bibliography{ref}
\bibliographystyle{plainnat}

\clearpage
\appendix
\section{Performance Evaluation Details} \label{sec:metric_details}

We employ various methods to quantify performance depending on the nature of the questions posed. All performance values are between 0 and 1, facilitating straightforward interpretation and aggregation across easy, medium, and hard questions through averaging.

\begin{itemize}
    \item \textbf{Q1, Q3, Q6:} These questions are evaluated based on correctness within a specified threshold. An answer provided by the language model (LLM) is deemed correct if the absolute difference between the LLM's answer and the ground truth answer is within 0.5\% of the ground truth value. If this condition is met, the LLM answer receives a performance value of 1; otherwise, it receives a value of 0. This approach accounts for floating-point precision issues and minor discrepancies arising from variations in coding logic in the use of PandasInterpreter and PythonInterpreter.
    \item \textbf{Q2, Q5:} The performance for these questions is assessed through set intersection. The ground truth answers consist of lists of values, and performance is calculated as the number of overlapping elements between the LLM answer and the ground truth answer divided by the total number of elements in the ground truth list. This metric effectively captures the degree of agreement between the LLM's output and the expected results.
    \item \textbf{Q4:} Evaluation for this question involves a boolean assessment to determine whether the LLM's answer appears within the list of all possible correct answers. If the LLM's answer matches any entry in the ground truth list, it is assigned a value of 1 (correct); if not, it receives a value of 0 (incorrect).
    \item \textbf{Q7:} This question is evaluated using the average $R^2$ score. It requires a series of numerical predictions from the LLM. By comparing these predictions to the ground truth values, we compute an $R^2$ score for each instance. The overall performance for all Q7-type questions is represented by the average of these $R^2$ scores.
    \item \textbf{Q8-Q12:} These questions involve binary classification tasks and are evaluated using the average bootstrap F1 score. Each task generates a single class prediction. We perform 1000 bootstrap sampling iterations with replacements on 50 versions of each question, generated from different instantiations of the question template. This process generates 1000 F1 scores. The average of these scores serves as a representation of performance for these binary classification tasks. We utilize F1 scores instead of accuracy to better address potential class imbalances in the data.
    \item \textbf{Q13:} This question is evaluated based on an exact match criterion. The performance metric is a boolean value indicating whether the LLM's answer matches exactly with the ground truth answer.
\end{itemize}

\section{Additional LIT Example} \label{sec:additional_example}
\begin{figure}
    \centering
    \includegraphics[width=0.6\linewidth]{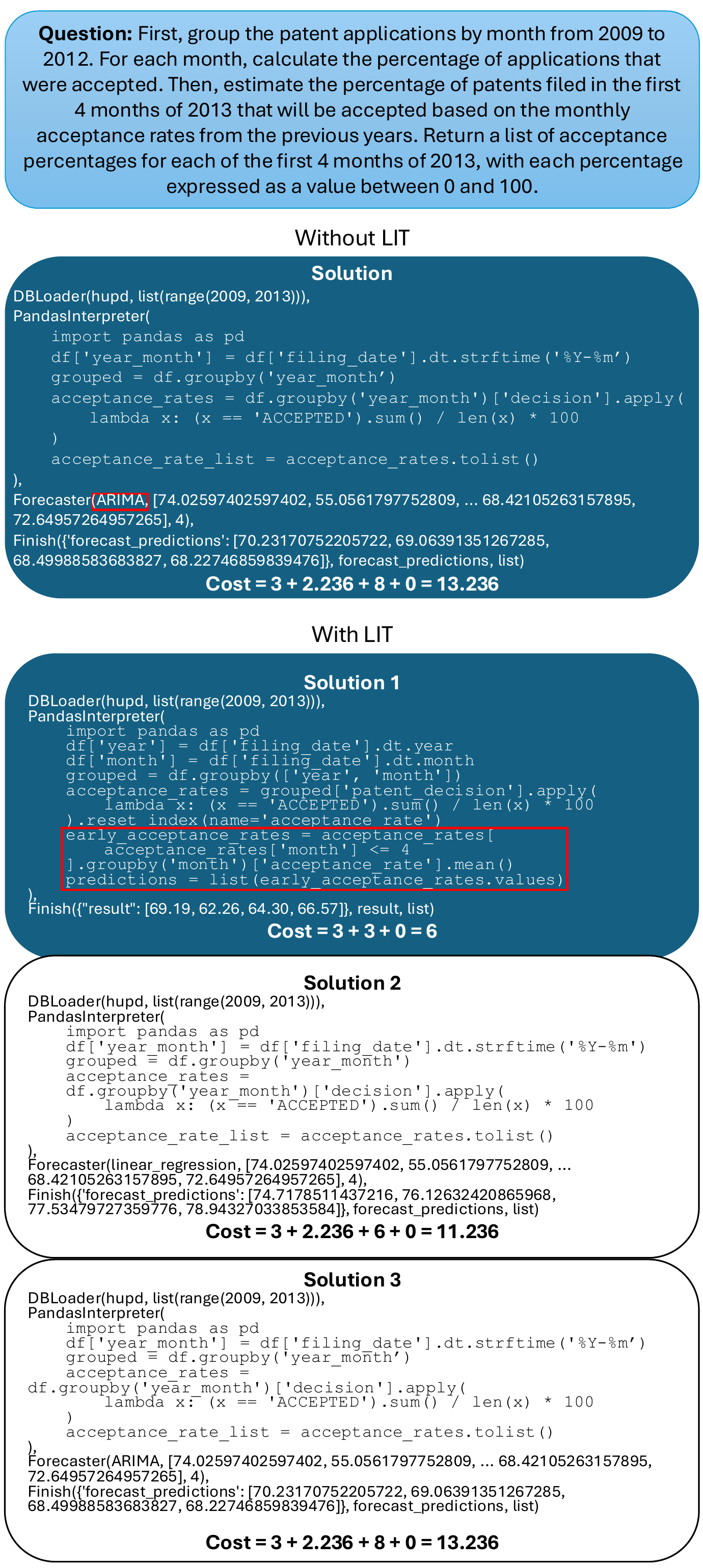}
    \caption{Solutions from an LLM to a question about future patent application acceptance, with and without LIT. When using LIT, the model generates multiple candidate solutions and selects the one with the best cost. As a result, the model with LIT chooses to use a simple average based on historical data rather than the ARIMA model used without LIT.}
    \label{fig:pandas_example}
\end{figure}

Figure \ref{fig:pandas_example} presents an illustrative case from Question 7 in which the model is asked to predict future patent application acceptance rates.
For this forecasting task, the black-box solution uses the PandasInterpreter to compute grouped acceptance rates from previous months and employs an ARIMA model-based Forecaster to predict acceptance rates for the subsequent four months. In contrast, LIT generates predictions directly using the PandasInterpreter. While both methods yield suboptimal predictions (with an $R^2$ score close to zero) the LIT approach is significantly easier to debug. This is because the PandasInterpreter's code argument is transparent, making it evident that the logic of relying only on the first four months of historical data fails to address the forecasting question accurately. Conversely, debugging the ARIMA model-based Forecaster is more challenging, as the issues could stem from the data's lack of stationarity or inappropriate hyperparameter settings. By employing LIT to select the solution, the debugging process becomes more straightforward.

\newpage
\section{Full Prompt Details} \label{sec:prompt_details}

\begin{figure*}
    \centering
    \includegraphics[width=1.0\linewidth]{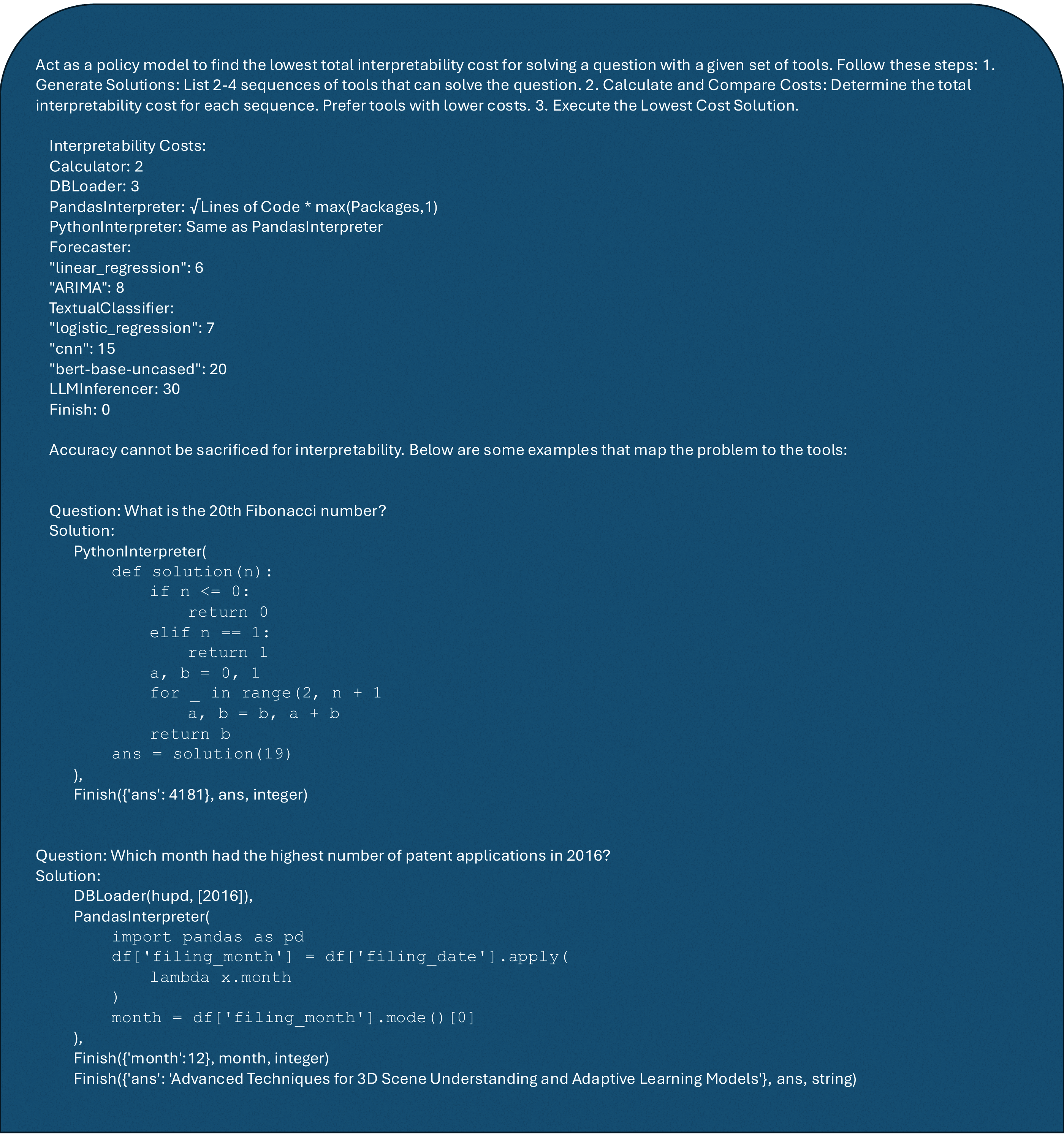}
    \vspace{1em}
    \caption{The full prompt presented to models in the LIT framework (continued in Figure \ref{fig:full-prompt-end}). Each LLM is presented with a set of costs, and instructed to form solutions that minimize these costs. A number of example solutions are provided.}
    \label{fig:full-prompt-start}
\end{figure*}

\begin{figure*}
    \centering
    \includegraphics[width=1.0\linewidth]{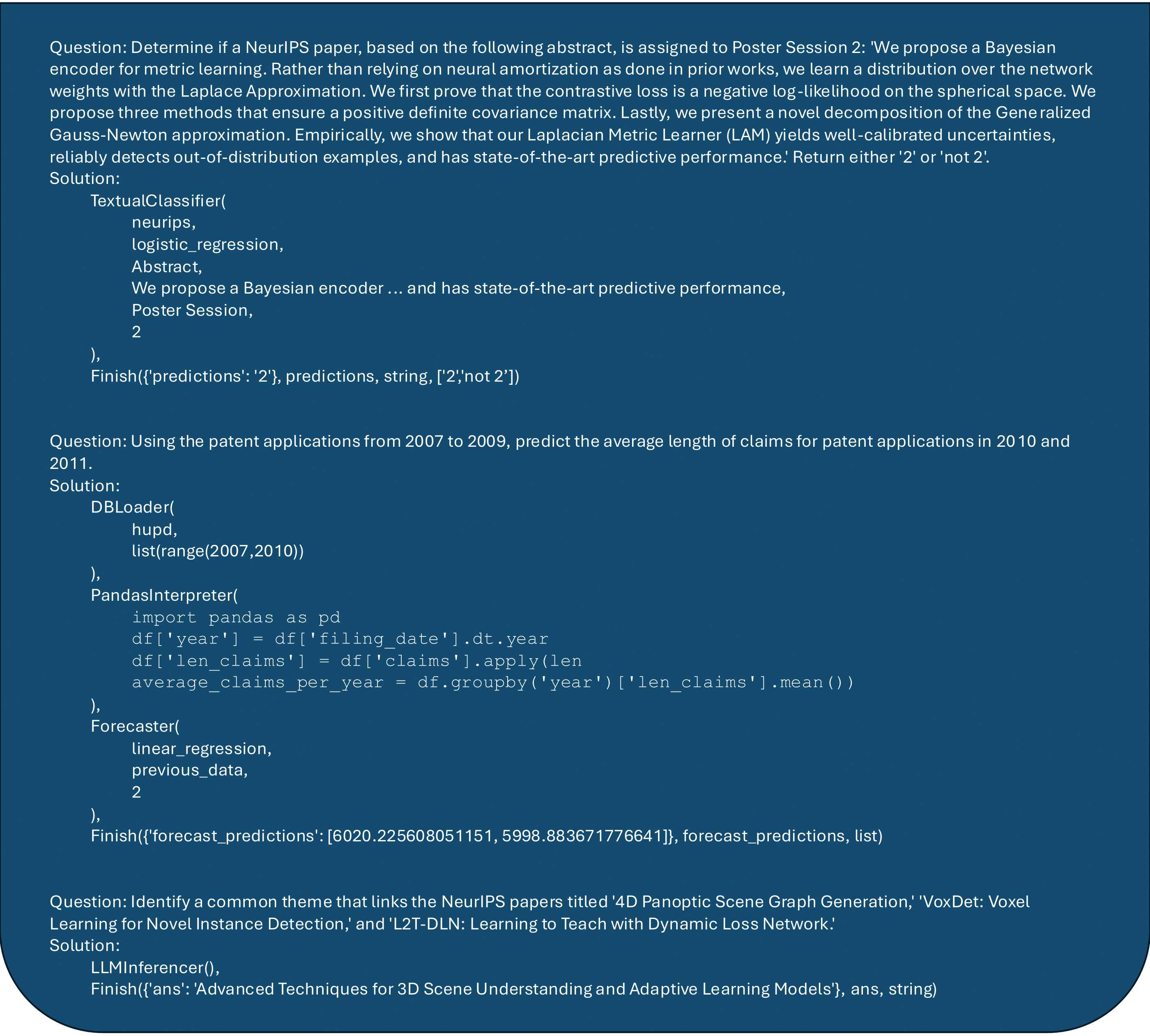}
    \vspace{1em}
    \caption{The full prompt presented to models in the LIT framework (continued from Figure \ref{fig:full-prompt-start}). Each LLM is presented with a set of costs, and instructed to form solutions that minimize these costs. A number of example solutions are provided.}
    \label{fig:full-prompt-end}
\end{figure*}

Here we describe the full context given to each model in the LIT framework. Figures \ref{fig:full-prompt-start} and \ref{fig:full-prompt-end} present the complete context. When using LIT, we instruct each LLM to generate multiple candidate solutions for each problem and select the most reliable/inspectable one according to our specified cost functions. We then provide five examples of successful solutions.

\vspace{1em}

\vspace{1em}

\clearpage
\begin{table*}
    \centering
    \resizebox{\textwidth}{!}{%
    \begin{tabular}{c|c|c|c|c}
        Tool & P & D & C & Cost=P+D+C \\
        \hline
        Calculator & 0 & 1 & 1 & 2 \\
        DBLoader & 0 & 2 & 1 & 3 \\
        PythonInterpreter & 0 & $\sqrt{\text{lines}} \times \max(\text{packages},1) \times 0.5$ & $\sqrt{\text{lines}} \times \max(\text{packages},1) \times 0.5$ & $\sqrt{\text{lines}} \times \max(\text{packages},1)$ \\
        PandasInterpreter & 0 & $\sqrt{\text{lines}} \times \max(\text{packages},1) \times 0.5$ & $\sqrt{\text{lines}} \times \max(\text{packages},1) \times 0.5$ & $\sqrt{\text{lines}} \times \max(\text{packages},1)$ \\
        Forecaster (linear regression) & 3 & 2 & 1 & 6 \\
        Forecaster (ARIMA) & 4 & 3 & 1 & 8 \\
        TextualClassifier (logistic regression) & 3 & 2 & 2 & 7 \\
        TextualClassifier (cnn) & 2 & 7 & 6 & 15 \\
        TextualClassifier (BERT) & 2 & 10 & 8 & 20 \\
        LLMInferencer & 1 & 15 & 14 & 30 \\
        Finish & 0 & 0 & 0 & 0 \\
    \end{tabular}%
    }
    \vspace{1em}
    \caption{Components of the cost in our experiments. Users of the LIT framework can choose their own tool costs based on their preferences. Cost components P, D, C refer to Robust Performance Across Inputs, Ease of Debugging, and Complexity of Arguments. In our experiments, we chose simple tools, such as the calculator, to have low P cost. PythonInterpreter and PandasInterpreter increase in cost when their results contain more lines or more packages.}
    \vspace{1em}
    \label{tab:cost_component}
\end{table*}
\vspace{1em}

\newpage
\section{Detailed Description of Question Bank} \label{sec:question_details}

Below are the 13 predefined question templates, each of which generates 100 variations by substituting the values inside the curly braces \{\}. Seven of these questions are related to the HUPD dataset, while six are related to the NeurIPS 2023 Papers Dataset, as noted in the brackets []. The questions are categorized by difficulty as follows: Q1-Q6 are ``easy,'' Q7-Q10 are ``medium,'' and Q11-Q13 are ``hard.'' 

\begin{itemize}
    \item Q1. [HUPD] What was the average time between the filing and issuance of patents from \{start\_year\} to \{end\_year\}? Return an integer representing the number of days by truncating the decimal part.
    \item Q2. [HUPD] What were the top \{\#\} \{IPCR/CPC categories\} with the highest number of accepted patents in \{year\}? Return them as a list of \{IPCR/CPC categories\}.
    \item Q3. [HUPD] How does the number of patent applications filed in \{year1\} compare proportionally to those filed in the \{year2\}? Return a number between 0 and 1. Please note that each row represents a patent application, and not all patent applications are assigned a patent number.
    \item Q4. [HUPD] What is the title of the patent filed between \{start\_year\} and \{end\_year\} that took the longest number of days between the filing date and the publication date?
    \item Q5. [NeurIPS] Who were the top \{\#\} authors with the most publications \{where the titles contain `Large Language Models'\} at NeurIPS?
    \item Q6. [NeurIPS] What proportion of NeurIPS papers have \{compare\} \{n\} authors? In the authors column of the database, each entry is a list, not a single string. Return a value between 0 and 1.
    \item Q7. [HUPD] First, group the patent applications by month from \{start\_year\} to 2012. For each month, calculate the percentage of applications that were accepted. Then, estimate the percentage of patents filed in the first \{n\} months of 2013 that will be accepted based on the monthly acceptance rates from the previous years. Return a list of acceptance percentages for each of the first \{n\} months of 2013, with each percentage expressed as a value between 0 and 100.
    \item Q8. [HUPD] For a patent application, which is not present in the database, with an abstract \{abstract\_content\}, predict whether it will get accepted. Return either `ACCEPTED' or `not ACCEPTED.'
    \item Q9. [NeurIPS] For a NeurIPS paper, which is not present in the database, with title \{title\_content\}, predict whether it belongs to \{topic\}? Return either `\{topic\}' or `not \{topic\}'.
    \item Q10. [NeurIPS] For a NeurIPS paper, which is not present in the database, with abstract \{abstract\_content\}, predict whether it will be accepted as an oral presentation? Return either `oral’ or `not oral’.
    \item Q11. [HUPD] Predict if the following two patents, which are not present in the database, belong to the same CPC category: \{title1\}, \{title2\}? Return `Yes' or `No'.
    \item Q12. [NeurIPS] Predict if this abstract-title pair, which is not present in the database, is from the same NeurIPS paper: Abstract: \{abstract\}. Title: \{title\}. Return `Yes' or `No'.
    \item Q13. [NeurIPS] Predict the best fit topic for the title of a NeurIPS paper, which is not present in the database: \{title\}. Options: \{topic1\}, \{topic2\}, \{topic3\}.
\end{itemize}

\section{Details of Model-Based Tools} \label{sec:modelbasedtools_details}
\begin{itemize}
\item \textbf{Forecaster:}
    \begin{itemize}
        \item \textbf{linear\_regression:} Utilizes the default settings of scikit-learn's linear regression model.
        \item \textbf{ARIMA:} Configured as a first-order autoregressive model (p=1), first-order differencing (d=1), and first-order moving average model (q=1).
    \end{itemize}
\item \textbf{TextualClassifier:}
    \begin{itemize}
        \item \textbf{bert-base-uncased, cnn, logistic\_regression:} Uses the default configurations of the respective models.
        \item \textbf{hupd:} Further trained on patents filed between 2004 and 2012. Related questions are based on the held-out set of patents filed between 2013 and 2018.
        \item \textbf{neurips:} Further trained on the first 3,000 papers from the NeurIPS 2023 Papers Dataset. Related questions are based on the held-out set of the last 590 papers in the dataset.
    \end{itemize}
\end{itemize}

\end{document}